\documentclass[]{interact}
\usepackage{tikz}
\usetikzlibrary{arrows}
\tikzstyle{var}=[circle,draw=black,fill=white,thin,minimum size=18pt,inner sep=0pt]
\tikzstyle{ivar}=[circle,draw=red,fill=white,thin,minimum size=18pt,inner sep=0pt]
\tikzstyle{avar}=[circle,draw=blue,fill=white,thin,minimum size=18pt,inner sep=0pt]
\tikzstyle{varh}=[circle,draw=gray,fill=white,thin,minimum size=18pt,inner sep=0pt,dashed]
\tikzstyle{arr}=[->,>=stealth',draw=black,thick]
\tikzstyle{arrh}=[->,>=stealth',draw=gray,thick,dashed]
\tikzstyle{biarr}=[<->,>=stealth',draw=black,fill=black,thick]
\tikzstyle{biarrh}=[<->,>=stealth',draw=gray,fill=gray,thick]
\usepackage{epstopdf}
\usepackage{subcaption}

\usepackage{natbib}
\bibpunct[, ]{(}{)}{;}{a}{}{,}
\renewcommand\bibfont{\fontsize{10}{12}\selectfont}

\defcitealias{biggs2021model}{Biggs and Sun, et al. 2021} 
\defcitealias{subramanian2022constrained}{Subramanian and Sun, et al. 2022} 
\defcitealias{CFX}{Tsiourvas and Sun, et al. 2024} 
\defcitealias{greenewald2021high}{Greenewald and Shanmugam, et al. 2021}

\theoremstyle{plain}

\theoremstyle{definition}

\theoremstyle{remark}

\begin{document}


\title{PresAIse, A Prescriptive AI Solution for Enterprises}


\author{
\name{Wei Sun\textsuperscript{a}\thanks{CONTACT AUTHOR: Wei Sun. Email: sunw@us.ibm.com}, Scott McFaddin\textsuperscript{a}, Linh Ha Tran\textsuperscript{b*}\thanks{*This work was done when the third author was an intern at IBM Research}, \\Shivaram Subramanian\textsuperscript{a},
Kristjan Greenewald\textsuperscript{a}, \\ Yeshi Tenzin\textsuperscript{a}, Zack Xue\textsuperscript{a}, Youssef Drissi\textsuperscript{a}, and Markus Ettl\textsuperscript{a}
 }
\affil{\textsuperscript{a}IBM Research; \textsuperscript{b}Rensselaer Polytechnic Institute}
}

\maketitle

\begin{abstract}
Prescriptive AI represents a transformative shift in decision-making, offering causal
insights and actionable recommendations. Despite its huge potential, enterprise adoption often 
faces several challenges. The first challenge is caused by the limitations of observational data for accurate causal
inference which is typically a prerequisite for good decision-making. The second pertains to the interpretability of recommendations, which is crucial
for enterprise decision-making settings. The third challenge is the silos between data scientists
and business users, hindering effective collaboration. This paper outlines an initiative
from IBM Research, aiming to address some of these challenges by offering a  suite of prescriptive AI solutions.
Leveraging insights from several research papers, the solution suite includes scalable causal
inference methods, interpretable decision-making approaches, and the integration of
large language models (LLMs) to bridge communication gaps via a conversation
agent. A proof-of-concept, PresAIse, demonstrates the solutions' potential by enabling
non-ML experts to interact with prescriptive AI models via a natural language interface, democratizing advanced
analytics for strategic decision-making.
\end{abstract}


\section{Introduction}

Prescriptive AI has emerged as a paradigm shift in the landscape of intelligent decision-making systems. 
Unlike descriptive analytics, which illuminates historical patterns, and predictive analytics, which anticipates future outcomes, prescriptive AI takes a leap forward by not only predicting potential outcomes but also recommending optimal courses of action. Despite artificial intelligence making huge strides in the last decade, there are several obstacles standing in the way of enterprise adoption of prescriptive AI.

\textit{Challenge 1:  Limitation of observational data for causal inference}

\noindent While an increasing number of individuals are becoming aware that correlation does not necessarily indicate causation and the latter is vital for decision making, enterprise data often consists of observational data, which is collected without the explicit manipulation of variables. Unlike experimental data, where researchers can control variables to establish causation, observational data which lacks randomization presents inherent limitations in establishing causal relationships due to factors such as  confounding effects.

\textit{Challenge 2:  Interpretability of prescriptive AI recommendations}

\noindent Traditional interpretability methods  focus on understanding the predictive power of AI models \citep{lundberg2017unified,ribeiro2016model}, where the goal is to reveal patterns and associations without explicitly addressing causation. In contrast, interpretability in prescriptive models used in decision-making settings  requires explaining the rationale behind the recommendations, the consideration of trade-offs, and the impact of interventions on desired outcomes.

\textit{Challenge 3: Silos between data scientists and business end users  }

\noindent Silos that exist in large enterprises can create significant barriers that hinder the smooth integration and effective adoption of AI technologies. These silos often result in a communication gap between business end users who understand the strategic goals of the organization and data scientists who possess the technical expertise. Business end users may not fully comprehend the technical intricacies of AI, while data scientists may struggle to convey the business value of their work. Ineffective AI adoption often results in wasted resources in terms of time and cost. Moreover, a lack of collaboration between the teams prevents the exploration of novel solutions and limits the organization's ability to reap the full benefits of AI.

In this paper, we describe an ongoing initiative at IBM Research that is aimed to address some of the aforementioned challenges that enterprises encounter when adopting prescriptive AI.   The primary objective of this initiative is to aggregate and synthesize valuable assets from multiple research papers in prescriptive AI from IBM research (i.e., \citetalias{greenewald2021high}; 
\citetalias{biggs2021model}; 
\citetalias{subramanian2022constrained}; \citealt{sun2023learning}; \citealt{OMT}; \citetalias{CFX}), creating a comprehensive and user-friendly suite of prescriptive AI solutions for enterprises. 
Here are some key functionalities that have been incorporated as part of the initiative.

\begin{itemize}
    \item With enterprise observation data, 
traditional causal inference approaches struggle to scale to real-world business use cases due to the number of variables and actions involved, the \textbf{sparsity-driven structure learning} approach proposed in \citetalias{greenewald2021high} is capable of scaling up to big data regimes, in some cases providing an exponential improvement in sample complexity, to facilitate accurate treatment effect estimation.

\item To aid interpretability in causal decision-making, one approach is to construct a \textbf{optimal prescriptive tree} where each path from the root to a leaf represents a policy. \citetalias{subramanian2022constrained} formulated this as  a novel mixed integer optimization problem and proposed a scalable algorithm to solve it via column generation. The output is human-readable rules which are characterized by the conditions present in each node along the path, making it accessible to non-technical stakeholders, such as business users and policymakers.

\item 
In the context of AI adoption for enterprises, we believe LLMs can play an important role in breaking the silos between business end users and data science groups. LLMs create Natural Language  Processing (NLP) interfaces which 
can pass the user's queries to downstream AI models responsible for prescriptive tasks, such as policy generation and  counterfactual predictions. The AI models generate responses or insights that are then presented back to the user in a human-readable format. We have created a proof-of-concept, named \emph{PresAIse} (pronounced `Pre-say-ice'), \textbf{an LLM agent with a NLP interface} which enables non-ML experts to interact with prescriptive AI models. Our goal is that by democratizing access to advanced analytics, we empower business users to explore, understand, and leverage the power of AI in driving strategic decisions.

\end{itemize}

Throughout the paper, we will focus on a use case of pricing premium seats for an airline as the running example to  provide a concrete and tangible illustration of the concepts discussed in the paper. The business end users in this case are pricing analysts who determine  pricing policy for a particular route that consists of an origin airport and a destination.

The rest of the paper is structured as follows: In Section \ref{sect_literature}, we will discuss related literature. We will briefly highlight two novel algorithms in the solution suite in Section \ref{sect_model}, which have been turned into API calls that can be queried via the conversation agent.  In Section \ref{sect_LLM}, we will review the architecture and key components in the LLM  agent. We will discuss future work and conclude the paper in Section \ref{sect_conclusion}. 

\section{Related art} \label{sect_literature}

Causal inference and causal decision-making are two closely related but distinct areas within the broader field of causal reasoning. While there has been a significant amount of research and literature focused on causal inference \citep{dudik2011doubly,athey2016recursive,wager2018estimation,kunzel2019metalearners,nie2021quasi} which primarily deals with the identification and estimation of treatment effect, the exploration of causal decision-making has received much less attention  \citep{fernandez2022causal}. This discrepancy highlights an intriguing gap in the understanding and application of causal knowledge in practical decision-making processes.

In recent years, there has been a stream of nascent research 
 focusing on learning interpretable optimal policy in the form of 
{prescriptive trees} for causal decision-making (\citetalias{biggs2021model}; \citealp{amram2022optimal,zhou2022offline}), since the tree structure is visually easy to understand. In contrast to standard decision trees for prediction tasks, prescriptive trees are constructed to provide policies that optimize a given objective (e.g., pricing strategies that maximize revenue, remedy action to minimize attribution). 
 Various tree-building algorithms have been proposed, e.g., the prescriptive tree can be constructed either greedily (\citetalias{biggs2021model}; \citealp{zhou2022offline}) or optimally using a mixed integer programming (MIP) formulation (\citealp{jo2021learning}; \citetalias{subramanian2022constrained}). Moreover, within the tree-constructing procedure, 
 the  prediction step may be explicitly decoupled from  the policy optimization (\citealp{amram2022optimal}; \citetalias{subramanian2022constrained}) or it can also be
embedded in the policy generation \citep{kallus2017recursive,bertsimas2019optimal,sun2023learning}. We want to point out that the current prescriptive tree algorithm supported in the LLM agent is proposed in \citetalias{subramanian2022constrained}, where we solve a constrained optimal prescriptive tree problem and  introduce a  novel path-based MIP formulation which identifies  optimal policy efficiently via column generation. The policy generated can be represented as a multiway-split tree which  is more interpretable and informative than a binary-split tree due to its shorter rules.

LLMs are increasingly being explored for their potential in simplifying and enhancing API  (Application Programming Interface) calling process \citep{qin2023toolllm,patil2023gorilla}. With their natural language understanding capabilities, LLMs can generate context-aware queries and parameters, allowing developers to interact with APIs using human-like language. LLMs are also increasingly finding applications in the field of Operations Research (OR). \cite{yang2023large}  leverages LLMs as optimizers, where the optimization task is described in natural language. Instead of using LLMs to replace optimizers, \cite{li2023large} use LLM in tandem with solvers where an LLM translates a human query into optimization code which is then fed into an optimization solver. 
This idea of using LLMs to query OR models is also explored in this paper. In particular, we explore using LLMs to query prescriptive AI models which combine both causal prediction and policy optimization. Moreover, we also explore using agents with memory to facilitate dynamic and context-aware interactions in an ongoing dialogue.


\section{Methodology} \label{sect_model}
Here we highlight the two foundational components of the prescriptive AI decision-making toolkit, for  causal inference and policy learning respectively.

\subsection{Scalable causal inference via structure learning}

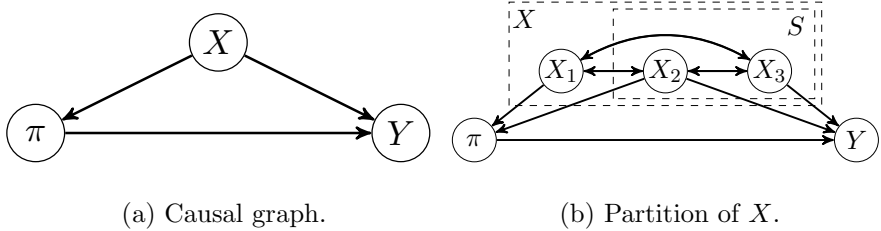
\begin{figure}[tb]
\centering
\begin{subfigure}{0.4\textwidth}
\begin{center}
\resizebox{\columnwidth}{!}{%
\begin{tikzpicture}
      \node[var] (Y) at (4,0) {$Y$};
      \node[var] (T) at (0,0) {$\pi$};
      \node[var] (X) at (2,1) {$X$};
      \draw[arr] (T) edge (Y);
      \draw[arr] (X) edge (T);
      \draw[arr] (X) edge (Y);
\end{tikzpicture}
}
\end{center}
\caption{Causal graph.}
\label{fig:graphX}
\end{subfigure}
\begin{subfigure}{0.4\textwidth}
\begin{center}
\resizebox{\columnwidth}{!}{%
\begin{tikzpicture}
      \node[var] (Y) at (5.5,0) {$Y$};
      \node[var] (T) at (0,0) {$\pi$};
      \node[var] (S) at (2.75,1) {$X_2$};
      \node[var] (Xp) at (1.25,1) {$X_1$};
      \node[var] (X3) at (4.25,1) {$X_3$};
      \draw[arr] (X3) edge (Y);
      \draw[arr] (X3) edge (S);
      \draw[arr] (S) edge (X3);
      \draw[arr] (Xp) edge[bend left] (X3);
      \draw[arr] (X3) edge[bend right] (Xp);
      \draw[arr] (T) edge (Y);
      \draw[arr] (S) edge (T);
      \draw[arr] (S) edge (Y);
      \draw[arr] (Xp) edge (T);
      \draw[arr] (Xp) edge (S);
      \draw[arr] (S) edge (Xp);
      \draw[dashed] (.5,.5) rectangle (5,2);
      \node[text width = .7cm] at (.9,1.75){$X$};
      \draw[dashed] (2,.6) rectangle (4.9,1.9);
      \node[text width = .7cm] at (4.85,1.65){$S$};
\end{tikzpicture}
}
\end{center}
\caption{Partition of $X$.  }
\label{fig:graph}
\end{subfigure}
\caption{Causal graph. $\pi$ is a discrete treatment, taking up to $q$ values, and $Y$ is a scalar outcome. $X$ is an observed set of $p$ covariates. }
\end{figure}

Consider the problem of estimating the treatment effect of a discrete-valued action $\pi$ on a univariate outcome $Y$ in the presence of covariates $X$, where the action variable $\pi$ can take $q = |\Pi|$ possible treatment configurations. We assume only observational data is available.\footnote{As opposed to inverventional data where actions were randomly assigned.} The causal graph for this setup is shown in Figure \ref{fig:graphX}.


One of the central issues of causal effect estimation is identifying features that are confounders and controlling for them. Let $Y_t(X)$ denote the counterfactual outcome associated when treatment $t$ is applied as an intervention given $X$.  
For any treatment $t$, let $Y_t \perp \pi |X$, i.e. the counterfactual outcome associated with any treatment $t$ is independent of the treatment choice in the observational data given $X$ (i.e. there are no hidden confounders). In this regime, causal effects can be estimated from observational data. Inverse propensity weighing, standardization and doubly robust estimation are standard techniques used \cite{guo2020survey,imbens2009recent}.

If $X \in \mathbb{R}^p$ is high dimensional (large $p$), however, the number of samples required to estimate the treatment effects  accurately becomes too large to be practical in many applications. This is at cross purposes with the need to avoid omitting hidden confounders - since these are not known a priori, features in $X$ should include as many possible confounding factors as possible. 
This makes handling the curse of dimensionality crucial for scaling causal methods to real world problems.

Consider a finer causal model given in Figure \ref{fig:graph}, where $X$ has been decomposed into the sets $X_1$, $X_2$ (confounders), and $X_3$ (predictors) based on their connections to $Y$ and $\pi$. 
Using the \emph{admissibility} formalism presented in Chapter 11 of \cite{pearl2009causality}, in \cite{greenewald2021high} we established that there is no \emph{additional} bias resulting from controlling for either $S$ only or $X_1 \cup X_2$ only instead of the full $X$ -- in other words, there will be no bias if $X$ includes all confounders.
The question then is which of these two sets to control for. 
One might imagine that when $X$ is high dimensional, the sparser of the two admissible subsets $S$ and $X_1 \cup X_2$ should be used.
It has been shown, however \citep{witte2020efficient} that effect estimates with $S$ will always be lower variance than estimates using $X_1 \cup X_2$ since $S$ includes all predictors $X_3$ of $Y$. 



Given observational samples of $X, Y, \pi$, our goal is thus to find the smallest subset $S$ containing all nodes in $X$ that have an edge pointing towards $Y$ in the graph. Since we have assumed that the outcome $Y$ does not have any edge pointing to $X$ or $\pi$, it is sufficient to use observational data to condition on $T=t$ and find the set of nodes $S_t$ in $X$ that have edges connecting to $Y$ in the undirected graph, and then take the union over $t$ as $S = \bigcup_{t=1}^q S_t$. 


In \cite{greenewald2021high}, we proposed identifying this set $S$ with a structure learning procedure. Applying a sparsity promoting regularizer $\rho_\lambda$ on the row 2-norms we have the following objective function:
\begin{align}
\hat{\theta} =& \arg\min_{ \|\theta\|_{1,2} \leq R} \left\{\sum_{j=1}^q\left[ \frac{1}{2}\theta_{:j}^T \frac{X_j^T X_j}{n} \theta_{:j}  - \frac{y_j^T X_j}{n}\theta_{:j}\right]+ \sum_{i=1}^p  \rho_\lambda\left(\|\theta_{i:}\|_{2}\right)\right\},
\label{eq:nonconvex}
\end{align}
The sparsity promoting regularizer sparsifies the 2-norms of the \emph{rows} of $\theta$ rather than its elements, ensuring that selected covariates are available for all actions $j$. 

\textbf{Theoretical Guarantees} In \citep{greenewald2021high}, we showed that for a specific class of nonconvex regularizers (e.g. the SCAD \cite{fan2001variable} and MCP \cite{zhang2010nearly} penalties), the true support set $S$ can be recovered with high probability given a number of samples $n$ only logarithmic in the ambient dimension $p$. In \cite{greenewald2021high} we provide easily satisfied conditions under which the solution to \eqref{eq:nonconvex} is unique and can be obtained using proximal gradient descent.

\textbf{Downstream Causal Effect Estimation} Using the set $S$ discovered as above, we can use any causal effect estimator from the literature, including IPW, doubly robust estimators, causal forests \cite{wager2018estimation}, etc. to obtain both average effect estimates as well as any needed counterfactual estimates.\footnote{Importantly, note that while the objective function \eqref{eq:nonconvex} implies a linear model on the outcome conditioned on the action, this does not restrict the downstream choice of linear or nonlinear estimators. Indeed, due to the nature of sparse regression, support set recovery is highly robust to nonlinearity in the underlying system (c.f. the broad success of Lasso based methods).}


\subsection{Interpretable policy via prescriptive trees}

With $N$ observational data samples $\{(x_i,\pi_i,y_i)\}_{i=1}^N$, where $x_i$ are features,  $ \pi_i$ refers to the action chosen from a discrete set $\Pi$,
and $y_i$ is the uncertain quantity of interest. In the airline pricing use case,  $x_i$ represents features related to the context of a ticket (such as the advance purchase period, departure day of the week, one-way or round-trip, class of service, etc.), $\pi_i$ is the price of a ticket, and $y_i$ indicates whether the sale was successful.

The objective of the counterfactual model is to provide the counterfactual estimates across all possible actions (e.g., prices) for each sample, $g_{i,\pi}$, where $\pi\in \Pi$ for all $i=1,\cdots, N$. For the pricing use case, we first learn a classifier $f(x,\pi)$ that maps input $x_i$ and $\pi_i$ to the binary outcome $y_i$. The function $f(x,\pi)$ is also known as the demand function, which represents the probability of purchase when given price $\pi$ and booking context features $x$. The counterfactual outcome we are interested in is the expected revenue, i.e., $g_{i,\pi}=\pi f(x_i,\pi)$.

\begin{figure}
\centering
  \includegraphics[width=.5\columnwidth]{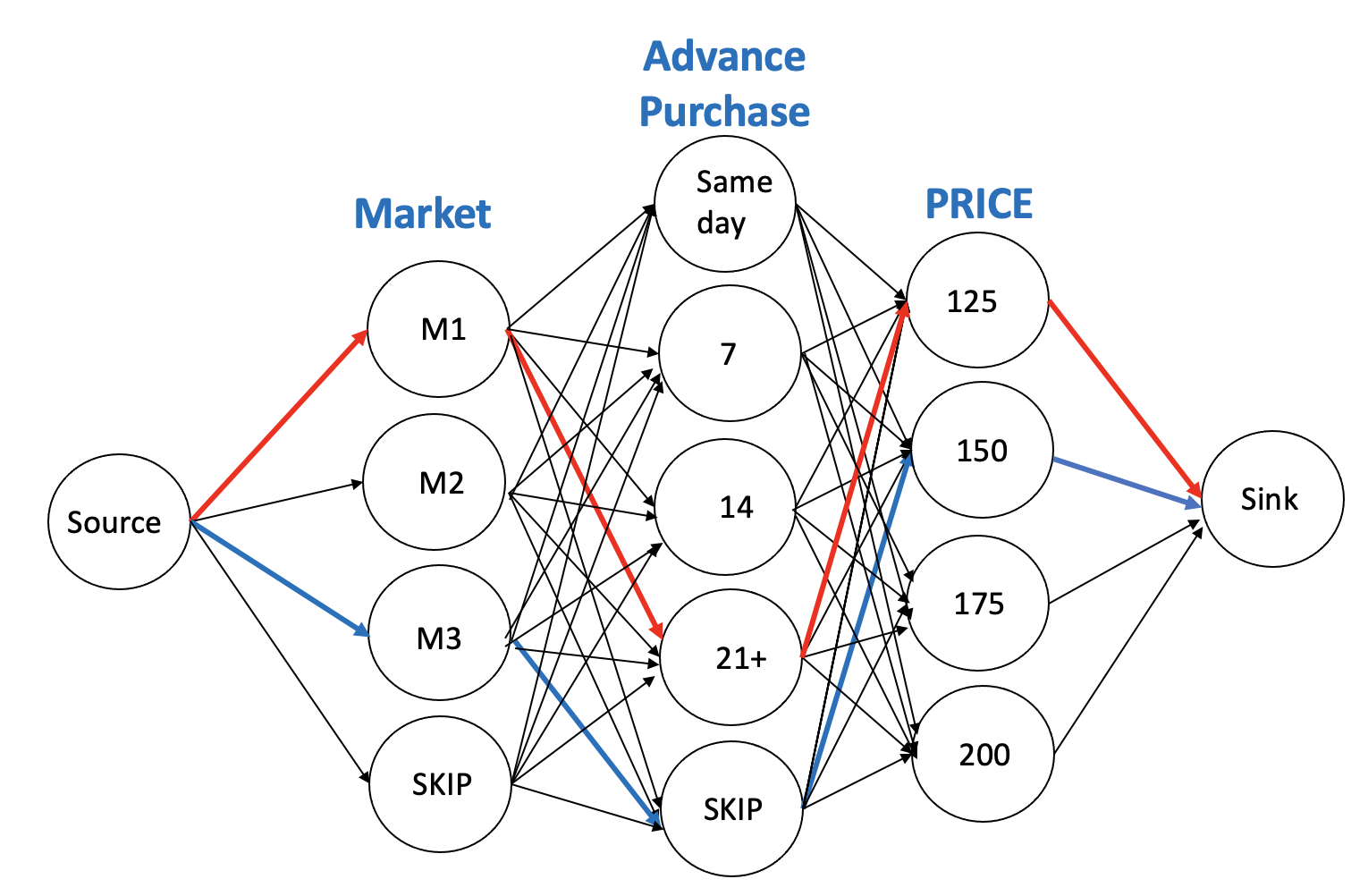}
  \caption{A feature graph with two features and the action (price). Two policies or decision rules are highlighted.}
  \label{fig:feature_space}
\end{figure}

The first step in constructing the optimal prescriptive tree (OPT) proposed in \citetalias{subramanian2022constrained} is  to define a feature graph that encompasses all possible combinations of input features and then to identify a subset of decision rules from the rule space. 
In a feature graph, each feature is represented by multiple nodes, each corresponding to a distinct feature value. Nodes of one feature are connected to nodes at the next level, and the graph also contains a source and a sink node. For each feature, with the exception of the action nodes $\pi$, there is a dummy node \emph{SKIP}. A decision rule is defined as a path $P\in\mathcal{P}$ from the source to the sink node in $G$. The rule space can be very large, as the number of possible rules is exponential in terms of the feature space. An example of a feature graph with two features (i.e., ``Market'' and ``Advance purchase window'') and the action nodes (``Price'') is shown in Fig~\ref{fig:feature_space}.

\begin{figure}
\centering
  \includegraphics[width=.7\columnwidth]{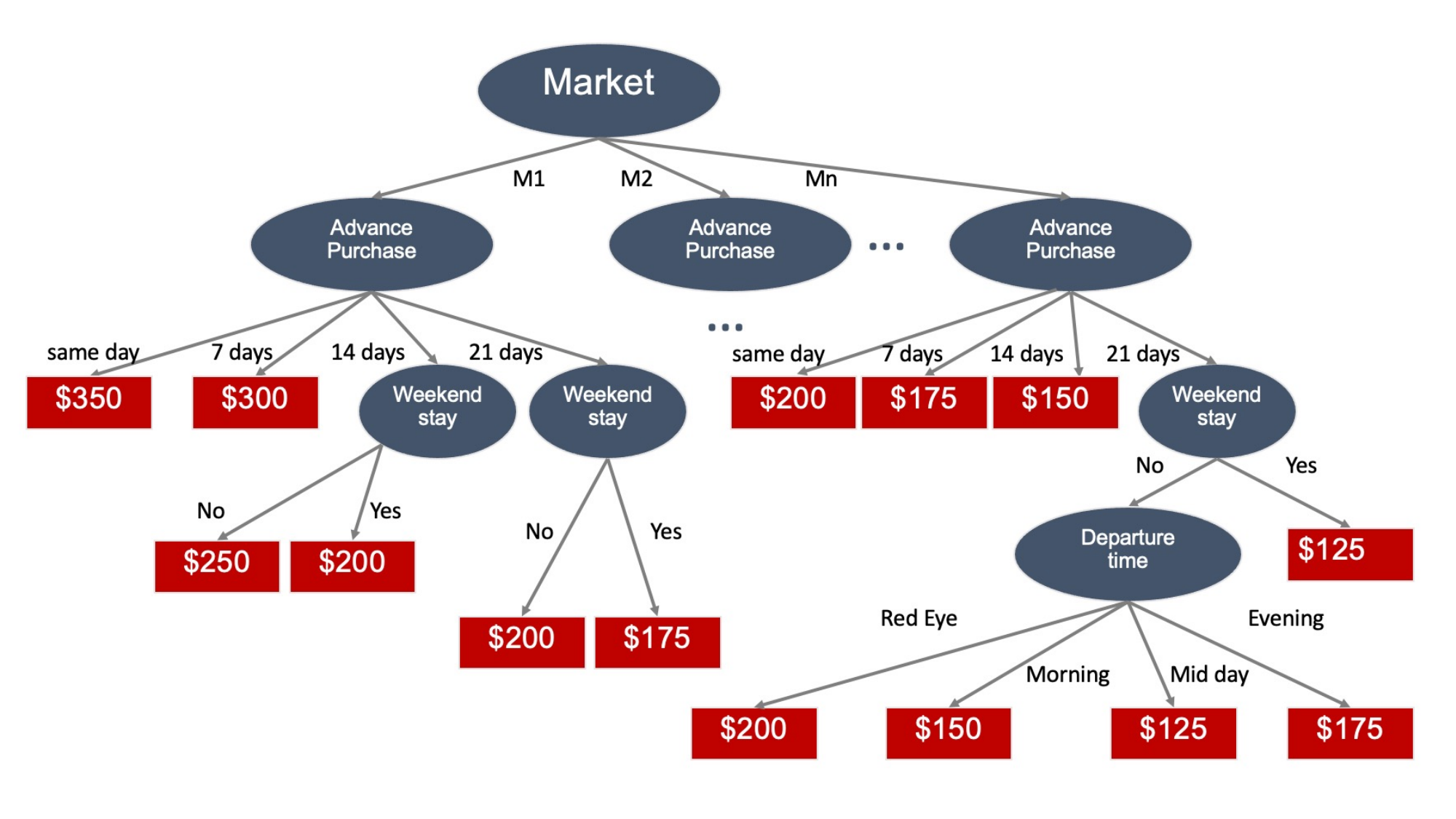}
  \caption{A prescriptive student tree for the airline pricing example}
  \label{fig:multiway_split}
\end{figure}

For each rule $j$ in the feature graph where $j=1,\cdots,M$ and $M=|P|$, define $r_j$ as the corresponding outcome in terms of the counterfactuals, i.e., $r_j = \sum_{i \in S_j} g_{i,p_j}$, where $S_j \in [N]$ is the subset of observations which fall into rule $j$ and $p_j\in \Pi$ is the action prescribed to rule $j$. The task of selecting a subset of $m$ rules from a feature graph with $M$ feasible paths can be formulated as MIP problem:

 \begin{align}\small
\quad\max\quad  \sum_{j=1}^M r_jz_j -\sum_{i=1}^N c_is_i \quad 
\textbf{s.t.} \quad  &\sum_{j=1}^M a_{ij}z_j + s_i = 1, \quad\forall i=1,\cdots,N\label{constraint_coverage}\\
&\sum_{j=1}^M z_j \leq m \label{capacity}\\
  &z_j\in\{1,0\}, \; \forall j=1,\cdots,M \nonumber ; \; s_i \geq 0, \; \forall i=1,\cdots,N \nonumber
\end{align}
where $a_{ij}=1$ if sample $i$ satisfies the conditions specified in rule $j$ and 0 otherwise. While a sample may fall into several rules, the set partitioning constraint (\ref{constraint_coverage}) along with the non-negative slack variables $s_i$ that are included with a sufficiently large penalty $c_i$ ensure that each sample is ultimately covered by exactly one rule.
The cardinality constraint (\ref{capacity}) ensures that at most $m$ rules are active in the optimal solution where $m$ is a user-defined input. The optimal solution of the set partitioning problem corresponds to a multiway-split tree with $m$ rules.
While  the set partitioning problem is NP-hard, \citetalias{subramanian2022constrained} propose a computationally efficient algorithm to solve the problem directly via dynamic column generation. An example of the prescriptive tree for the airline pricing use case is shown in Figure \ref{fig:multiway_split}.

\section{PresAIse, a prescriptive AI agent driven by LLMs} \label{sect_LLM}

PresAIse (pronounced `Pre-say-ice'), is a word play of ``prescriptive'' and ``AI''.  The goal of the system is to empower business end users to harness the benefits of AI without requiring specialized AI knowledge, by incorporating a LLM-powered NLP interface. 
\subsection{Overview}

\begin{figure}
\centering
  \includegraphics[width=.8\columnwidth]{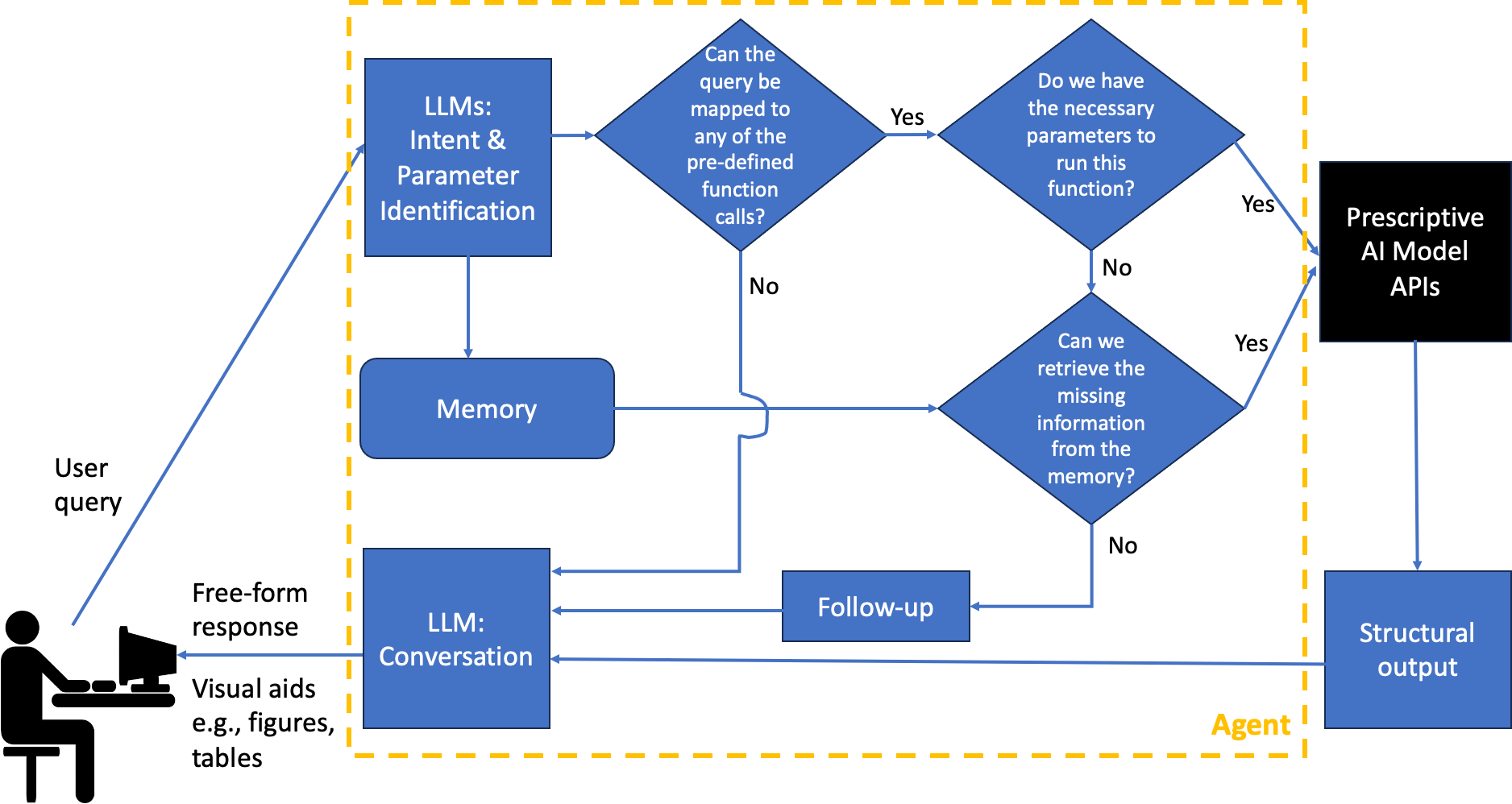}
  \caption{The framework for PresAIse, the prescriptive AI agent }
  \label{fig:flowchart}
\end{figure}

The framework of PresAIse, depicted in Figure \ref{fig:flowchart}, consists of four key entities: (1) LLMs, (2) tools in terms of API function calls from downstream prescriptive AI models, (3) agent and (4) memory. When a user inputs a query, an agent powered by LLMs determines whether one of the pre-defined API calls needs to be made based on the input. Under the hood, 
 LLMs perform several NLP tasks including intent classification as well as slot filling. More specifically, the intent of a query is mapped to one of the pre-defined API calls, while slot filling identifies various parameters or arguments to be used in API calls. When an LLM identifies an API and/or parameters, these values are written into a customized memory module.

Based on the output from the LLMs, there are four possible scenarios the agent can proceed: 1) the agent determines that the query is indicating a specific API call and necessary input arguments are present, then the  function call will be executed; 2) the agent determines that the user is asking for an API  call, but certain arguments are missing. In this case, the agent will first go into the memory to see if this information can be retrieved from the earlier conversations. If so, the agent will proceed to call the specific API. 3) On the other hand, if the agent is unable to locate the necessary information, she will ask a follow-up question for that information. 4) The agent determines that the user query is not related to any of the pre-defined API calls, the user query is fed into a dialog LLM to produce a response.

The system utilizes LangChain \citep{langchain}, an open-source software library for developing applications powered by language models.
We now provide more details for each of the components. 


\subsection{Back-end}

\subsubsection{Open-source LLMs}

PresAIse is built entirely with open-source LLMs. We have made a conscious decision to not use proprietary models such as ChatGPT, partly due to license/IP limitations, and partly because we believe for many enterprise applications, transparency, cost efficiency, and flexibility in customization of open-source models yields more benefits in the long run.

While fine-tuning allows training specifically for the target task, which can result in better performance on that particular task compared to a generic pre-trained model, it requires substantial computational resources, including powerful GPUs or TPUs. Moreover, fine-tuning approaches also require separate datasets for training. Highly specialized domains  require domain-specific language, terminology, and context. Acquiring or creating a sufficiently large and representative training dataset for fine-tuning in these domains can be challenging. To circumvent these obstacles, we apply
in-context learning, where the LLM performs a task just by conditioning on input-output examples, without optimizing any parameters. 

Through our experiments, we observe that instead of utilizing a single LLM for various NLP tasks on hand (i.e., API identification, slot filling, dialog), separate models that are dedicated to individual tasks achieve better performance. This is not surprising when a single model is used for multiple tasks, the context provided through prompts may become large and complex. This can lead to contextual confusion, where the model may struggle to discern the relevant details for each specific task within the lengthy prompt. Separating models avoids this confusion and ensures clarity. 
There are also additional benefits of using separate LLMs. For instance, managing separate models is more scalable, especially when new tasks are introduced or modifications are needed since changes to one model do not affect others. Moroever, task-specific models with in-context learning are often more interpretable. Users and developers can better understand how the model makes decisions in the context of a specific task, enhancing transparency and interpretability compared to a single, multi-task model.

We utilize Flan-T5 XXL \citep{flanT5} for dialog generation and several  LLMs including IBM Watsonx Granite models for the remaining NLP tasks including identifying API calls and input arguments.
We initialized all LLMs with a temperature of 0 except the model for dialog.  A low temperature is useful when using tools as it decreases the amount of “randomness” or “creativity” in the generated text of the LLMs, which is ideal for encouraging it to follow strict instructions — as required for tool usage.

For each LLM, we apply few-shot prompting for in-context learning, where we append the instruction with several domain-specific examples. 
Instructions for the LLMs  generating dialog and API classifications are shown in Table~\ref{tab:prompt-instruction}. In the prompt for the dialog LLM, we also utilize role prompting technique where the context is given in the form of a role. Table~\ref{tab:few_shot_examples} provides some examples that are used to teach LLMs to identify function calls as well as input arguments. 

While prompt engineering which involves crafting specific prompts and/or examples to elicit desired responses from language models is known to be time-consuming, the challenges and complexities amplify  significantly
when applied to domain-specific applications. This is because such applications often involve specialized terminology, jargon, or context-specific language, crafting effective prompts demands domain expertise. Domain-specific queries or prompts may often contain ambiguities that need to be carefully addressed to balance specificity with adaptability and ensure that prompts are not overly rigid or ambiguous. Continuous collaboration with end users and domain experts is essential to iteratively refine and optimize prompts for effective model performance in such applications, which we leave as future work.

\begin{table}[ht]
    \centering
    \footnotesize
    \begin{tabular}{p{0.12\linewidth} | p{0.85\linewidth}}
Dialog & You are a friendly and cheery AI agent, your name is PresAIse, pronounced `Pre-say-ice'. Your role is to support airline pricing analysts in determining pricing policies. You analyze each user's query to determine their intent and map it to one of the underlying AI functions. You then execute the function and retrieve the results. Your purpose is to assist business users who are not computer scientists in making better decisions by utilizing the power of AI.
    At the same time, we are looking for ways to further enhance AI policies by incorporating users' domain knowledge and insights that aren't in the training data.
    We believe that the hybrid team you are forming with pricing analysts will outperform AI or human teams alone.
    
    Key functions that are currently supported include
    generate a set of optimized pricing policy, evaluate the KPIs in terms of revenue, conversion, and predict counterfactual scenarios.
    When a user query can be mapped to one of the existing functionalities with necessary parameters, reply with enthusiasm that you are happy to assist the user and you are working on the query. 
    When you do not understand the query or when the intent is not mapped one of the functionalities, ask for more clarification on the user's intent. 
     \\\hline
      Intent classification  & Classify command as one of the following API calls. If the intent is unclear, output ``Unknown". 
      \newline ``RUN\_OPT'' produces the optimized pricing policy for a given market.
     \newline ``CF\_PRICE\_BOUND" evaluates ``what if'' scenarios by predicting the counterfactual outcome in terms of the expected revenue when prices are adjusted.
     \newline ``SHOW\_BASE\_POLICY" shows the current or historical pricing policy for a given market. 
     \newline ``KPI\_REVENUE" returns the revenue for a given pricing policy for a given market. 
     \newline ``KPI\_CONVERSION" returns the conversion for a given pricing policy for a given market. 
     \newline $\cdots$
     \\\hline
    \end{tabular}
    \caption{Instructions used in the prompt for dialog and API classification LLMs. }
    \label{tab:prompt-instruction}
\end{table}

\begin{table}[ht]
    \centering
    \footnotesize
    \begin{tabular}{p{0.5\linewidth} | p{0.4\linewidth}}
Command: Can you show me the optimized pricing policy for Detroit to Los Angeles route? & function\_call: RUN\_OPT
            \newline origin-destination: DTW-LAX
     \\\hline
       command: Can we set the minimum price from \$200 to \$250, and re-optimize the pricing policy with no more than 10 rules? &
            function\_call: RUN\_OPT
            \newline price\_bound: 250-Unknown
            \newline cardinality: 10 \\\hline
        command: What if we lower the maximum price to \$550? &
        function\_call: CF\_PRICE\_BOUND
            \newline price\_bound: Unknown-550\\\hline
           Command: What is the conversion rate for flights departing JFK?&
            function\_call: KPI\_CONVERSION
            \newline origin-destination: JKF-Unknown
      \end{tabular}
    \caption{Sample examples used in intent classification and slot filling LLMs. }
    \label{tab:few_shot_examples}
\end{table}

\subsubsection{Tools}

In PresAIse, the tools are a set of APIs connected to downstream models that business end users can query. Some of the key function calls are shown in the instructions for the intent classification model in Table~\ref{tab:prompt-instruction}. There are standard business intelligence (BI) reporting functionalities, such as retrieving the current pricing policies and corresponding KPIs in revenue and conversion.

The  most critical function call is \textsc{run\_OPT}, which produces the optimized pricing policy for a given route. This function call involves causal feature selection, training the counterfactual outcome model as well as performing the optimization model to produce the optimal prescriptive tree (OPT) as described in Section~\ref{sect_model}.

Another important functionality is \textsc{CF\_price\_bound} which allows a user to ask the system ``what if'' questions and evaluate these counterfactual outcomes with prices different from the model's recommendation. We view such functionality as a critical step towards bringing human-in-the-loop. Firstly, these queries contribute to transparency and explainability as they enhance users' understanding of how the AI system operates and responds to inputs changes. Morever, AI models are typically trained with a specific objective. This singular focus might not fully capture the complexity and multifaceted nature of real-life decision-making scenarios. For instance, the ``optimality'' of the AI model is defined with respect to maximize the expected revenue.  During our engagement with the airline pricing analysts, we discover that there are many other objectives pricing analysts take into account when setting prices, eg., reserving a certain number of premium seats for free upgrades for frequent fliers so as to maintain customer loyalty and satisfaction. Allowing humans to pose ``what if" questions acknowledges that real-life decisions are often characterized by ambiguity and the need to balance conflicting considerations. Lastly, users often bring domain-specific expertise and external knowledge that may not have been explicitly captured in the training data of the AI model. Allowing users to pose ``what if" scenarios recognizes the invaluable insights and context they can provide, enriching the decision-making process.

\subsubsection{Agent}

Agents are components of the LangChain framework that are designed to interact with the real world. They are used to automate tasks and can perform a variety of tasks.
LLMs are  used as a reasoning engine to determine an agent's action.  In PresAIse, we construct  a custom multi-action agent who has access to tools described in the previous section.
Based on the output of LLMs, the agent  decides the next action to take, including executing one of the API calls, or asking a follow-up question for missing input.  Before asking the question, the agent will first attempt to look into the memory component which we will discuss next. 

\subsubsection{Memory}

By default, LLMs are stateless, that is, every incoming query is processed independently of other interactions. The memory allows a LLM to remember previous interactions with the user and enables coherent conversation. 
One of the most common types of memory involves returning entire past chat history. While storing everything gives the LLM the maximum amount of information and the process  is simple and intuitive, more tokens mean slowing response times and higher costs. Moreover, LLMs can only process a limited
number of tokens which is a strict constraint that developers need to find workarounds for. Hence, some memory only “remembers” the previous $k$ human-AI interactions where $k$ is a user input. To avoid excessive token usage,
we construct a customized memory system that only keeps specific contexts in terms of the recent API calls and input arguments to prevent memory overload and improve execution efficiency.

More specifically, the memory system maintains a JSON object that consists of  function call and all the input arguments that are being used by various functions. Upon model initialization,  the values are set to \textsc{Unknown}.
 The memory system supports two actions: writing and reading. After the LLMs parse  a user query, function call and/or parameters that are identified will be written into memory. While the agent  has access to the memory to retrieve input argument, but she is only allowed to read from the memory but no writing ability. For instance, in PresAIse, \textsc{origin-destination} is a necessary input argument used by all APIs currently supported. Once a pricing analyst identifies the market, future queries without mentioned \textsc{origin-destination} can be processed as the LLM agent retrieves this information from the memory.

\subsection{Front-end user experience}
\label{subsect_user_experience_principles}

The front-end UI is a user-friendly and interactive platform designed to seamlessly connect with the back-end system described in the previous section, facilitating intuitive communication and data visualization for a business end users. The UI comprises two main components: a chat interface and a high fidelity dashboard as would be commonly used in business intelligence (BI)
and analytics settings.

\begin{figure}
    \centering
    \includegraphics[width=\textwidth]{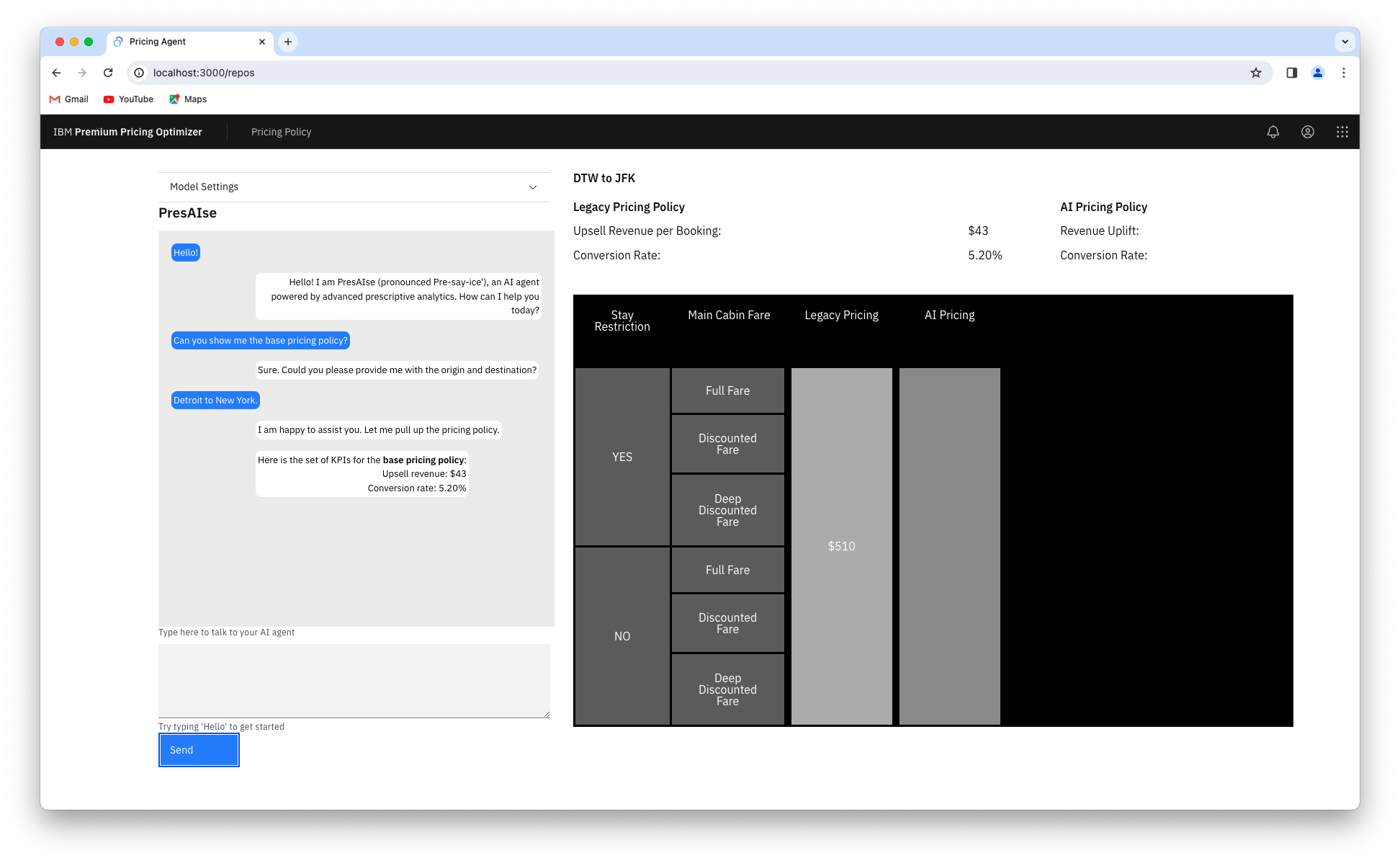}
    \caption{Query current pricing policy}
    \label{fig:intro-screen-shot}
\end{figure}

\begin{figure}
    \centering
    \includegraphics[width=\textwidth]{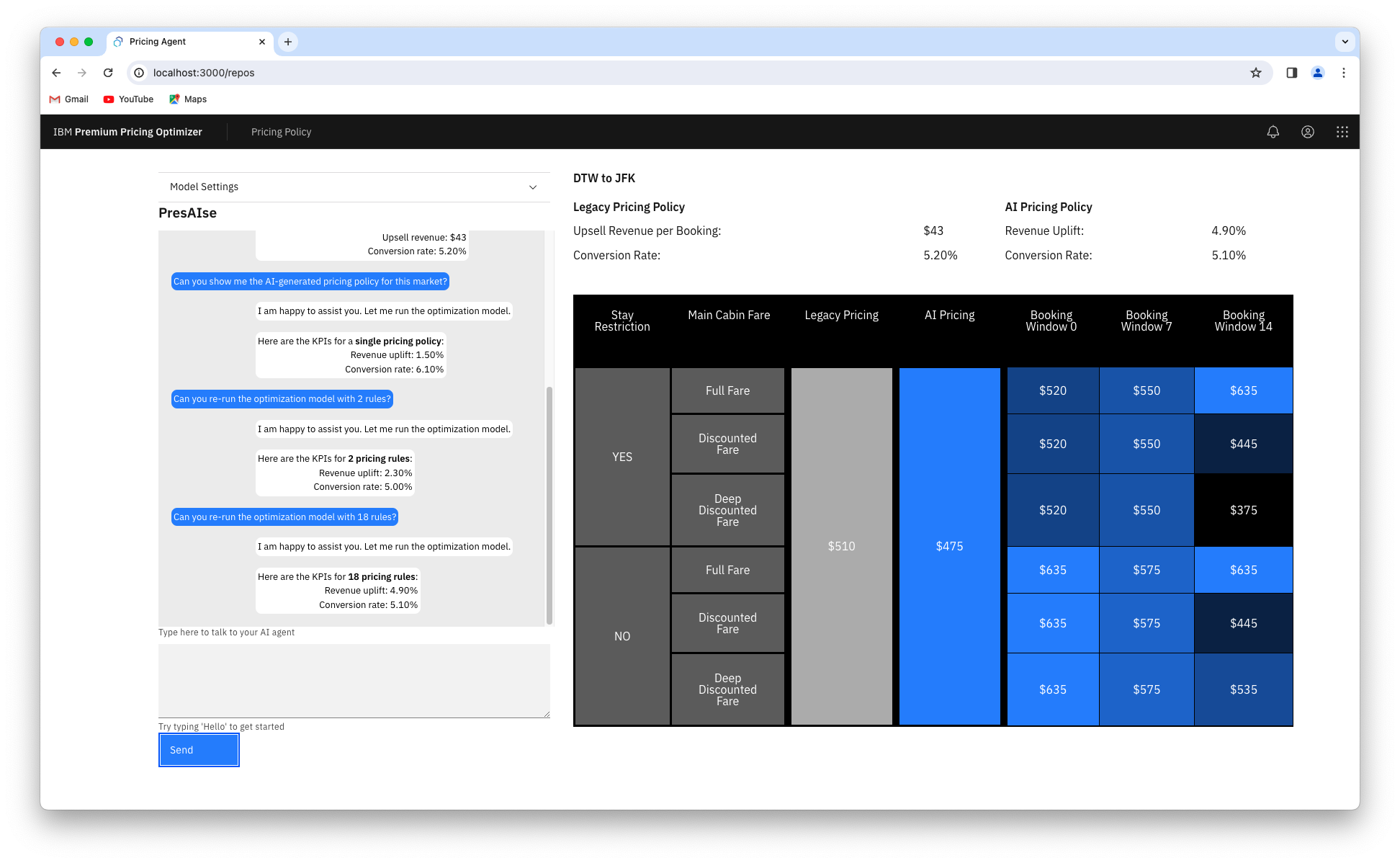}
    \caption{Query AI generated pricing policy}
    \label{fig:18-rule-screen-shot}
\end{figure}

Figure \ref{fig:intro-screen-shot} shows the screenshot when an user asks for the current pricing policy for DTW-JFK market via the \textsc{SHOW\_BASE\_POLICY} API call. Figure \ref{fig:18-rule-screen-shot} depicts when an user asks for a fully differentiated optimal pricing policy for the same market via executing the \textsc{RUN\_OPT} API call. We  refer readers to the appendix for additional scenarios.

\section{Conclusion}\label{sect_conclusion}
As the demand for data-driven decision-making continues to grow, the tools and methods presented in this paper offer some solutions for enterprises seeking to navigate the challenges of observational data, enhance interpretability, and extend the benefits of AI to a broader spectrum of users. By combining causal inference methodologies with user-friendly interfaces, our suite of tools strives to make advanced analytics an integral and accessible component of the decision-making landscape in modern enterprises.

\textbf{Ongoing work}
Firstly, we plan to add additional functionalities as tools for the LLM agent. For instance, we intend to include constraint identification which allows a business end user to specify additional constraints for the optimization model. We are also planning to include  covariate selection as an separate tool, e.g., the user may wish to understand which covariates have been selected, and conversationally request that certain covariates be added or removed (e.g. if a covariate was selected that in retrospect should not have been considered a possible confounder).  We also plan to include functionalities such as counterfactual explanation which helps to identifies the minimum changes needed to achieve a desirable outcome specified by an end user. 
Secondly, our work currently utilizes the few-shot in-context learning where the model learns new task by observing only a few query-response examples. Recent studies show that parameter-efficient fine-tuning, i.e a small set of parameters are trained to enable a model to perform the new task, offers better accuracy as well as dramatically lower computational costs than few-shot learning and standard fine-tuning \citep{NEURIPS2022_0cde695b,gu2022ppt}. 
Finally, in our current system, the agent only has access to the back-end prescriptive AI toolkit. The agent knowledge can be improved by connecting them to external database source and tools, which can be done via Retrieval Augmented Generation (RAG) \citep{lewis2021retrievalaugmented}. 


\section*{Disclosure statement}
\noindent The authors declare that there is no conflict of interest with respect to the research, authorship, and publication of this article.

\section*{Data availability statement}
The data that support the findings of this study are available from the corresponding author, W. S., upon reasonable request.

\bibliographystyle{tfcse}
\bibliography{prescriptive}

\end{document}